\pdfoutput = 1
\documentclass[twocolumn]{article}

\usepackage{times}
\usepackage[numbers]{natbib}
\usepackage[english]{babel}
\usepackage{blindtext}
\usepackage{graphicx}
\usepackage{amsmath, amsthm, amssymb, bbm, bm}
\usepackage{enumerate}
\usepackage{float}      
\usepackage{subcaption}  
\usepackage{wrapfig}
\usepackage[margin=0cm]{caption}
\usepackage[titletoc, toc]{appendix}
\usepackage{tabularx}

\usepackage{multirow}
\usepackage{hhline}
\usepackage{makecell}
\usepackage{placeins}  

\usepackage[x11names, usenames, dvipsnames, svgnames, table]{xcolor}
\definecolor{firebrick}{rgb}{.698,.133,.133}
\definecolor{mybluelight}{rgb}{0.9, 0.9, 1.}
\definecolor{myorangelight}{rgb}{1., 0.9, 0.9}

\usepackage[utf8]{inputenc} 
\usepackage[T1]{fontenc}    
\usepackage{url}            
\usepackage{booktabs, colortbl}       
\usepackage{amsfonts}       
\usepackage{nicefrac}       
\usepackage{microtype}      

\usepackage{csquotes}
\usepackage{latexsym}

\usepackage{pifont}
\newcommand{\cmark}{\ding{51}}%
\newcommand{\xmark}{\ding{55}}%

\usepackage[boxruled, vlined, linesnumbered]{algorithm2e}
\SetAlFnt{\small}
\SetAlCapFnt{\small}
\SetAlCapNameFnt{\small}
\usepackage{algorithmic}
\algsetup{linenosize=\tiny}

\let\oldnl\nl
\newcommand{\nonl}{\renewcommand{\nl}{\let\nl\oldnl}}

\usepackage{paralist}

\usepackage{xspace}
\usepackage{soul}
\usepackage{dsfont}
\usepackage{stmaryrd}
\usepackage[textwidth=15mm]{todonotes}
\usepackage{dirtytalk}
\usepackage{pbox}
\usepackage{cprotect}

\usepackage{verbatim}
\usepackage{textcomp}
\usepackage[normalem]{ulem}

\usepackage{mathtools}
\usepackage{etextools}
\usepackage[inline]{enumitem}

\usepackage[colorlinks=true,allcolors=firebrick,bookmarks=false]{hyperref}

\definecolor{darkergreen}{RGB}{21, 152, 56}
\definecolor{red2}{RGB}{252, 54, 65}
\definecolor{Gray}{gray}{0.85}
\newcolumntype{g}{>{\columncolor{Gray}}c}

\let\OLDthebibliography\thebibliography
\renewcommand\thebibliography[1]{
  \OLDthebibliography{#1}
  \setlength{\parskip}{0pt}
  \setlength{\itemsep}{0pt plus 0.3ex}
}

\theoremstyle{definition}

\DeclarePairedDelimiterX{\divx}[2]{(}{)}{%
  #1\;\delimsize\|\;#2%
}

\newcommand*{\etal}{\emph{et al}\@\xspace}

\usepackage{style}

\makeatletter
\@namedef{ver@everyshi.sty}{}
\newcommand{\removelatexerror}{\let\@latex@error\@gobble}

\newcommand\blfootnote[1]{%
  \begingroup
  \renewcommand\thefootnote{}\footnote{#1}%
  \addtocounter{footnote}{-1}%
  \endgroup
}

\title{Textualized and Feature-based Models for Compound Multimodal Emotion \\Recognition in the Wild}

\renewcommand\footnotemark{}

\author{
  Nicolas~Richet$^{1}$,
  ~Soufiane~Belharbi$^{1}$,
  ~Haseeb~Aslam$^{1}$,
  ~Meike~Emilie~Schadt$^{1}$,
  ~Manuela~González-González$^{2,3}$,\\
  ~\textbf{Gustave~Cortal}$^{4,6}$,
  ~\textbf{Alessandro~Lameiras~Koerich}$^{1}$,
  ~\textbf{Marco~Pedersoli}$^{1}$, 
  ~\textbf{Alain~Finkel}$^{4,5}$,
  ~\textbf{Simon~Bacon}$^{2,3}$, \\ and
  ~\textbf{Eric~Granger}$^{1}$\\
 	$^1$ LIVIA, ILLS,  Dept. of Systems Engineering, ETS Montreal, Canada \\
	$^2$ Dept. of Health, Kinesiology \& Applied Physiology, Concordia University, Montreal, Canada\\
  $^3$ Montreal Behavioural Medicine Centre, CIUSSS Nord-de-l’Ile-de-Montréal, Canada\\
  $^4$ Université Paris-Saclay, CNRS, ENS Paris-Saclay, LMF, 91190, Gif-sur-Yvette, France\\
  $^5$ Institut Universitaire de France, France \\
  $^6$ Université Paris-Saclay, CNRS, LISN, 91400, Orsay, France \\
{\tt\footnotesize \textcolor{black}{\{nicolas.richet.1,muhammad-haseeb.aslam.1\}@ens.etsmtl.ca} } \\
{\tt\footnotesize \textcolor{black}{\{soufiane.belharbi,eric.granger\}@etsmtl.ca} }
}

\newcommand{\ignore}[1]{}

\newcommand\meld{\texttt{MELD}\xspace}
\newcommand\cexprdb{\texttt{C-EXPR-DB}\xspace}  
\newcommand\cexprdbours{\texttt{C-EXPR-DB}$^*$\xspace} 

\begin{document}
\maketitle\thispagestyle{fancy}

\maketitle
\rhead{\color{gray} \small Richet et al. \;  [ECCVw 2024]}

\begin{abstract}
Systems for multimodal emotion recognition (ER) are commonly trained to extract features from different modalities (e.g., visual, audio, and textual) that are combined to predict individual basic emotions. However, compound emotions often occur in real-world scenarios, and the uncertainty of recognizing such complex emotions over diverse modalities is challenging for feature-based models.
As an alternative, emerging large language models (LLMs) like BERT and LLaMA can rely on explicit non-verbal cues that may be translated from different non-textual modalities (e.g., audio and visual) into text. Textualization of modalities augments data with emotional cues to help the LLM encode the interconnections between all modalities in a shared text space. In such text-based models, prior knowledge of ER tasks is leveraged to textualize relevant non-verbal cues such as audio tone from vocal expressions, and action unit intensity from facial expressions. Since the pre-trained weights are publicly available for many LLMs, training on large-scale datasets is unnecessary, allowing to fine-tune for downstream tasks such as compound ER (CER).
This paper compares the potential of text- and feature-based approaches for compound multimodal ER in videos. Experiments were conducted on the challenging \cexprdb dataset in the wild for CER, and contrasted with results on the \meld dataset for basic ER. Our results indicate that multimodal textualization provides lower accuracy than feature-based models on \cexprdb, where text transcripts are captured in the wild. However, higher accuracy can be achieved when the video data has rich transcripts.
Our code is available at: \href{https://github.com/nicolas-richet/feature-vs-text-compound-emotion}{github.com/nicolas-richet/feature-vs-text-compound-emotion}\blfootnote{\copyright 2025 The Author(s), under exclusive license to Springer Nature Switzerland AG.  \href{https://doi.org/10.1007/978-3-031-91581-9_5}{https://doi.org/10.1007/978-3-031-91581-9\_5}}.
  
\end{abstract}


\begin{figure*}[!htp]
\centering
  \centering
  \includegraphics[width=\linewidth]{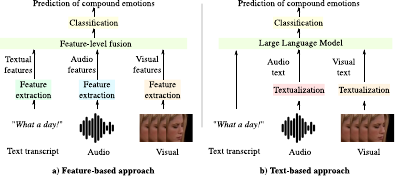}
  \caption[Caption]{Models for compound multimodal ER in videos. 
  (a) In the \emph{feature-based approach}, a dedicated feature extractor produces embeddings for each input modality. A feature-level fusion module then combines embeddings from all different modalities to produce joint feature representations for classification.
  (b) In the \emph{text-based approach}, textual descriptions are extracted for nonverbal modalities, such as audio and visual. These texts are combined with verbal cues (i.e., text transcripts) and fed to an LLM as a joint textual representation for classification.
  }
  \label{fig:promo}
\end{figure*}

\section{Introduction} 
\label{sec:intro}

Emotion recognition (ER) plays a critical role in human behavior analysis, human-computer interaction, and affective computing~\cite{ezzameli23,vijayaraghavan24}. Research in ER mainly focuses on recognizing the seven basic emotions -- anger, surprise, disgust, enjoyment, fear, sadness, and contempt~\cite{belharbi24-fer-aus, Matsumoto1992}. Recently, there has been growing interest in recognizing complex emotions commonly occurring in real-world scenarios, such as compound emotions, where a mixture of emotions is exhibited~\cite{kollias23}. ER is more challenging for complex emotions because they are often ambiguous and subtle, and can be easily confused with basic emotions. Despite the availability of multiple modalities that can potentially help to recognize these complex emotions~\cite{zhang24}, they can introduce additional uncertainty and conflict~\cite{ji23,wang22}.

Multimodal information, e.g., faces, voice, and text extracted from videos, has been used extensively to develop robust ER models~\cite{aslam24,vijayaraghavan24,Waligora-abaw-24}. Convolutional and transformer-based backbones are commonly trained to extract discriminant features from each modality. These include vision backbones such as ResNet~\cite{heZRS16} and audio backbones such as VGGish~\cite{hershey17}.  
A fusion model is required to combine features from verbal (spoken text) and nonverbal cues (visual and audio), thereby producing contextualized features to predict accurate emotion classes~\cite{liang24}. Multimodal learning allows for building complex joint feature representations that can achieve high accuracy for ER. This feature-based approach has driven much progress in ER~\cite{vijayaraghavan24}. Indeed, it requires simple and minimal emotion-class annotations -- typically enough to allow these models to learn to \emph{automatically} extract and combine relevant features from different modalities for prediction.  
However, using only a single emotion class for CER in real-world scenarios and without any other guidance, such as output supervision or input cues, is challenging, especially using videos captured in the wild~\cite{kollias24-6}.

Recently, another multimodal learning approach called TextMI has been proposed for sentiment analysis~\cite{hasan23} in videos. In contrast with the feature-based approach, input modalities like audio and visual are textualized. Based on prior knowledge of the task, the authors propose extracting nonverbal cues deemed relevant to the task in the form of a descriptive text. This can include a textual description of action unit (AU) intensity from visual~\cite{ekman1978,friesen1978} and the tone of the audio. This conversion of modalities to text could be seen as an \emph{expert-based data augmentation} with emotion-related cues used as inputs for the model during training and evaluation. This augmentation provides the models with direct guidance and emotional context to learn the task at hand. Since all modalities are formatted as text, a language model is required to better understand the interconnections between words and modalities and their relations to emotions. However, state-of-the-art language models are typically large, and training them requires a considerable amount of text data, which is not always available. The recent surge of large language models (LLMs)~\cite{xin23} made their use possible~\cite{hasan23}. Powerful LLMs such as BERT~\cite{Devlin19} and LLaMA~\cite{touvron23} have been pre-trained, and their weights have been made public, allowing us to fine-tune these models for downstream tasks~\cite{raj24}. Their application in multimodal ER provides a simple method for multimodal fusion, and results shown in~\cite{hasan23} are promising. Multimodal textualization remains largely unexplored in the literature. In a very recent work~\cite{cheng24}, modality descriptions are employed to model output supervision for emotional reasoning and recognition. This differs from TextMI~\cite{hasan23}, which uses these cues as inputs. In this paper, we analyze text-based approaches like TextMI for multimodal CER.  

\begin{figure*}[!ht]
\centering
  \centering
  \includegraphics[width=0.8\textwidth]{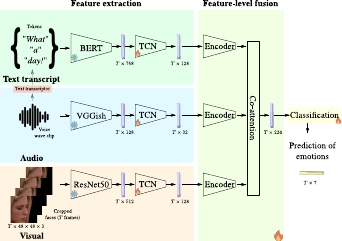}
  \caption[Caption]{A common feature-based approach used for multimodal CER.}
  \label{fig:feature}
\end{figure*}

This paper focuses on the following question: \emph{how does textualized modeling perform against feature-based modeling for CER in videos?} As shown in Fig.~\ref{fig:promo}(a), feature-based methods employ different backbone models to automatically extract features from each modality. Feature-level fusion models also allow us to automatically learn joint feature representations for ER, although combining diverse modalities over videos remains a challenge~\cite{liang24}. The textualized approach~\cite{hasan23} simplifies fusion since all nonverbal modalities (visual and audio) are converted into a single textual modality (see Fig.~\ref{fig:promo}(b)). However, a bottleneck of the approach is that textualizing nonverbal modalities requires the choice of textual descriptions. For instance, the same audio segment can be textualized into the high/low valence/arousal spectrum, or the vocal intonation can be textualized. Similarly, for the facial modality, the choice of AUs and the granularity and context window are some of the design choices that can heavily affect the overall performance. The manual selection and design of textual descriptions require the intervention of domain experts to construct relevant nonverbal cues, making them application-dependent. This is similar to the contrast between using learned versus handcrafted features and the challenges of the latter.
 
In this paper, we compare the performance of state-of-the-art deep learning models that follow standard feature-based vs. text-based modeling approaches. An extensive set of experiments was conducted on the challenging \cexprdb video dataset for CER in the context of the \emph{7th Workshop and Competition on Affective Behavior Analysis in-the-Wild (ABAW)}~\cite{kollias24,kollias24-7,kollias24-6,kollias23-2,kollias23abaw,kollias22abaw,kollias21,kollias21-1,kollias20}. To further assess the benefits of using a textualized approach in basic ER, experiments were also conducted on the \meld video dataset.

\section{Feature-based Modeling} 
\label{sec:feature-based}

This approach extracts features from audio (vocal) and video (facial) modalities and text transcripts for multimodal CER in videos. Feature embedding is combined for feature-level classification (see Fig.~\ref{fig:feature}). The rest of this section provides more details on this approach. 

\noindent \textbf{Feature Representation. }  
In ER applications, the de facto strategy to leverage different modalities extracted from videos is to extract their features~\cite{aslam24,kuhnke20,schoneveld21,Waligora-abaw-24,zong23}.  
These modalities typically include vision and audio. Textual modality, such as audio transcripts, is also included when available. Other ER applications such as pain estimation leverage bio-signals such as physiological modalities~\cite{Waligora-abaw-24,werner14}: electrodermal activity (EDA), electromyograph (EMG), and electrocardiogram (ECG). The general motivation behind combining these modalities is to leverage their complementary information over a video sequence.

Each modality typically employs a dedicated pre-trained feature extractor, which can be pre-trained on different large-scale datasets. In addition, public weights pre-trained on related datasets can be employed. For visual modality, ResNet~\cite{heZRS16} backbone is commonly used, which is followed by a module that leverages temporal information such as temporal convolutional network (TCN)~\cite{bai18}. 3D models such as R3D-CNN~\cite{tran18} can better leverage spatio-temporal dependency between frames directly at feature extraction. For audio modality, a variety of public pre-trained feature extractors are available, such as VGGish~\cite{hershey17}, Wav2Vec 2.0~\cite{wagner23}, and HuBERT~\cite{hsu21}. In addition, traditional audio features can be easily computed, such as spectrograms and MFCCs~\cite{xu04}. Multiple text feature extractors are available for the text modality, such as BERT~\cite{Devlin19} and RoBERTa~\cite{liu19}. Feature extractors are typically kept frozen while the subsequent modules are fine-tuned to avoid expensive computational costs.

\begin{figure*}[!t]
\centering
  \centering
  \includegraphics[width=\textwidth]{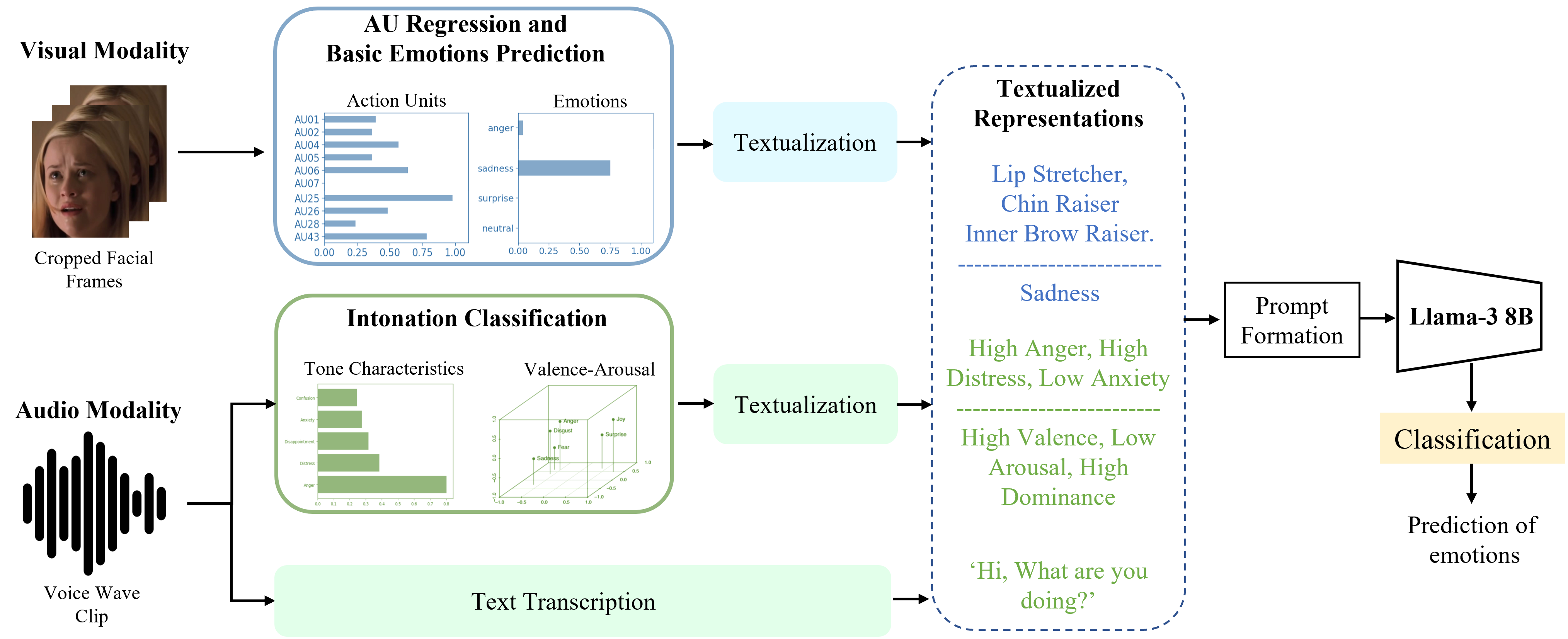}
  \caption[Caption]{A common text-based approach used for multimodal CER, where non-verbal modalities are textualized.}
  \label{fig:textualize}
\end{figure*}

\noindent \textbf{Feature-Level Fusion. }  
A bottleneck in feature-based models is fusion (Fig.~\ref{fig:promo}). Different methods rely on temporal models to combine features from over a video, like LSTMs~\cite{deng21,schoneveld21,nguyen21}, or rely on simple concatenation~\cite{kuhnke20,zhi21}. Recent works focus more on self- and cross-modal attention and transformers~\cite{vaswani17} to perform attention-based fusion~\cite{le23,lu23,praveen21,tran22,Waligora-abaw-24,zhang22,zhou23}. This has shown promising results as they can capture inter- and intra-modality relationships. Note that aligning modalities in a video setup is challenging as well. For example, what part of audio or text could be assigned to a single frame remains unclear. Usually, an empirical sliding window is employed to align other modalities with a frame.

This paper follows the recent work~\cite{zhang22} to experiment with a feature-based approach for the ER task (Fig.~\ref{fig:feature}). Their method achieved good results for continuous ER (valence/arousal) in videos in a recent ABAW challenge. In our case, it was adapted to perform emotion classification in videos as well. In particular, we use ResNet50~\cite{heZRS16}, pre-trained on MS-CELEB1M~\cite{guo16} and FER$^+$~\cite{barsoum16} datasets for visual modality. For audio modality, we employ VGGish~\cite{hershey17}, while BERT~\cite{Devlin19} is used over text modality. Temporal information is further exploited by using temporal convolutional network (TCN)~\cite{bai18} after each feature extractor. We employ the co-attention block~\cite{zhang22} to attend to features from different modalities. This builds a single embedding per frame while leveraging a contextual window. The per-frame feature is then fed to a classifier head to predict emotions.

\section{Text-based Modeling} 
\label{sec:textualization}

This approach extracts textual descriptions from audio (vocal) and visual (facial) modalities for multimodal CER in videos, and combines them with text transcripts. This joint textual description is processed by an LLM for classification (see Fig.~\ref{fig:textualize}). The rest of this section provides more details on this approach.

\noindent \textbf{Audio and Visual Text Description.}  
The API of Hume Inc.\footnote{\scriptsize\href{https://www.hume.ai/}{https://www.hume.ai}} is employed over a sliding window to analyze the tone of the audio. Their model is trained on millions of human interactions\footnote{\scriptsize It is based on an empathic large language model (eLLM), combining language modeling and text-to-speech.}. The API scores each tone characteristic, allowing us to sort and select the top 10. Examples of tone characteristics include confusion, anxiety, disappointment, distress, and even basic emotions. The name of each tone characteristic and a "Low" or "High" prefix, determined using a threshold on the score, are used to describe the tone textually. We also use a fine-tuned Wav2Vec 2.0 model~\cite{wagner23}\footnote{\scriptsize\href{https://huggingface.co/audeering/wav2vec2-large-robust-12-ft-emotion-msp-dim}{huggingface.co/audeering/wav2vec2-large-robust-12-ft-emotion-msp-dim}} to predict scores for arousal, valence and dominance for each audio clip. Those scores are then categorized as "Low" or "High" using a threshold. The textual description produced from audio concatenates all tone and arousal-valence-dominance textual descriptions. 

Face cropping and alignment are first performed at each frame using RetinaFace~\cite{dong19}. Py-feat library~\cite{cheong23} is then used to extract action units (AUs) intensity~\cite{ekman1978,friesen1978} along with basic emotion probabilities. AUs codebook~\cite{belharbi24-fer-aus} maps each facial expression to a set of action units. For instance, the expression "Happy" is associated with "AU6" (Cheek raise), "AU12" (Lip corner puller), and "AU25" (lips part). Typically, a set of these facial AUs activate at once. Over a sequence of frames, we select the maximum intensity of each action unit over time. Then, we use a threshold to determine which AUs are activated. In this case, the text description will be the concatenation of the name of each selected action unit as depicted in Fig.~\ref{fig:textualize}. The same procedure is repeated with basic emotions, where the top 3 emotions are selected, and their textual name is used as a description.

\noindent \textbf{Combination of Transcripts with Audio and Visual Texts.}  
Some multimodal datasets like MELD \cite{Poria19} and CMU-MOSEI \cite{zadeh18} provide text transcriptions. We use Whisper-large-v2~\cite{radford23} to generate transcripts when they are unavailable. Once all textual descriptions are acquired, we combine them into a single prompt and feed it to an LLM such as LLaMA-3~\cite{meta-llama3-24}, in our case. Our used prompts follow this template: 
\\
\noindent{\scriptsize\texttt{
"
\\
Speech transcription of the video : [transcription]; 
\\
Facial Action Units activated during the video : [visual\_AU\_text]; 
\\
Emotions predicted from visual modality: [visual\_emotions\_text]; 
\\
Characteristics of the prosody : [tone\_description]; 
\\
Audio emotional state : [arousal-valence-dominance\_text]
\\
"}}. 
\\
\noindent A fully connected layer (FC) follows LLaMA-3 to perform classification. Both the LLM and the FC layer are fine-tuned using ground-truth class labels.

\section{Results and Discussion} 
\label{sec:exps}

This paper compares feature- and text-based approaches for multimodal CER in videos in the wild. For a fair comparison, our experimental setup is constrained to be as similar as possible for the two approaches. This section provides the experimental methodology, results, and discussion. We also include results on basic emotions in videos.

\subsection{Experimental Methodology}
\label{subsec:exp-meth}

\noindent\textbf{(1) Datasets.} Two video-based datasets for emotion recognition are used: \cexprdb (compound emotion), and \meld (basic emotion).

\noindent a) \emph{\cexprdb}~\cite{kollias23}: It is composed of 56 videos taken from 7th ABAW CER Challenge test set. The full \cexprdb~\cite{kollias23} contains 400 videos with 200,000 frames in total. However, we do not have access to the 400, but only to the 56 test videos of the challenge. Each frame has an annotation with twelve compound emotions. For the 7th ABAW Challenge, only seven compound emotions are considered from the original twelve, which are: Fearfully Surprised, Happily Surprised, Sadly Surprised, Disgustedly Surprised, Angrily Surprised, Sadly Fearful, and Sadly Angry. In this work, these 56 videos used for the test are referred to as \cexprdb. The challenge organizers provide it without annotation. To perform experiments earlier than the challenge deadline, we annotated these 56 videos by our internal expert team. Each video may have different parts where there are compound emotions. The annotation is done at frame level with the same seven compound emotions of the challenge in addition to the class "Other" that represents any other emotion, compound or basic, that is different from the considered seven ones. Once annotated, we cut each original video into segments using the annotation timeline. Each segment contains only one compound emotion. We obtained 125 segments. We refer to this dataset as \cexprdbours. Its class distribution is presented in Table.\ref{tab:our-clips}. We split \cexprdbours into 5-cross validation. Performance is reported on the validation set of each fold. The evaluation is done only on compound emotions while discarding the class "Other". However, the training may or may not include this extra class. More details about our annotation are provided later.

{
\setlength{\tabcolsep}{3pt}
\renewcommand{\arraystretch}{1.1}
\begin{table*}[!h]
\centering
\resizebox{.5\textwidth}{!}{%
\centering
\begin{tabular}{lc||c|rc}
\hline
\textbf{Compound class}   &&  \textbf{Number of segments}  && \textbf{Total duration (secs)}\\
\hline \hline
Angrily Surprised         && \ \ \ 8              && \ 53.92 \\
Sadly Angry               && \ 16                && 100.60 \\
Fearfully Surprised       && \ 24                && 98.85 \\
Happily Surprised         && \ 15                && 150.57 \\
Sadly Fearful             && \ 19                && 107.56 \\
Disgustedly Surprised     && \ 10                && \ 82.45 \\
Sadly Surprised           && \ 13                && 182.17 \\
Other$^*$                 && \ 20                && \ 71.00 \\
\hline
\textbf{Total}            && 125               && 847.15 \\
\hline
\end{tabular}
}
\caption{Class distribution and duration of our internally labeled \cexprdbours dataset from the 56 videos of the 7th ABAW CER Challenge taken from \cexprdb dataset.
\\
$^*$: In some of our experiments, the class "Other" has been used only for training. Only the seven compound emotions are considered for evaluation.}
\label{tab:our-clips} 
\end{table*}
}

{
\setlength{\tabcolsep}{3pt}
\renewcommand{\arraystretch}{1.1}
\begin{table*}[!t]
\centering
\resizebox{\linewidth}{!}{%
\centering
\begin{tabularx}{\linewidth}{lc|X}
\hline
\textbf{Code}   &&  \textbf{Definition} \\
\hline \hline
Fearfully Surprised       && Experiencing a sudden shock or surprise accompanied by fear. 
                             This might happen if something unexpected occurs that also 
                             appears threatening or dangerous. Must include at least one 
                             cue related to fear and a cue related to surprise, either 
                             occurring simultaneously or closely following each other.\\
\hline
Happily Surprised         && Experiencing a sudden and unexpected event that brings joy
                             or happiness. This type of surprise is pleasant and delightful. 
                             Must include at least one cue related to happiness and at least
                             one cue related to surprise. They should occur simultaneously or 
                             closely following each other.\\
\hline
Sadly Surprised           && Experiencing a sudden and unexpected event that brings sadness or 
                             disappointment. This type of surprise is unpleasant and upsetting. 
                             Must include at least one cue related to sadness and at least one 
                             cue related to surprise. They should occur simultaneously or closely 
                             following each other.\\
\hline
Disgustedly Surprised     && Experiencing a sudden shock or surprise that also causes feelings of 
                             disgust or revulsion. This might happen if something unexpected occurs 
                             that is also repulsive. Must include at least one cue related to disgust
                             and at least one cue related to surprise. They should occur simultaneously
                             or closely following each other.\\
\hline
Angrily Surprised         && Experiencing a sudden shock or surprise that provokes anger. This might 
                             happen if something unexpected occurs that is also infuriating. Must 
                             include at least one cue related to anger and at least one cue related 
                             to surprise. They should occur simultaneously or closely following each
                             other.\\
\hline
Sadly Fearful             && Feeling both sadness and fear simultaneously. This can occur when facing
                             a situation that is both threatening and sorrowful. Must include at least 
                             one cue related to sadness and at least one cue related to fear. They should
                             occur simultaneously or closely following each other.\\
\hline
Sadly Angry               && Feeling both sadness and anger simultaneously. This can happen when dealing 
                             with a situation that evokes both sorrow and frustration or rage. Must include
                             at least one cue related to sadness and at least one cue related to anger. 
                             They should occur simultaneously or closely following each other.\\
\hline
\end{tabularx}}
\caption{Codebook used by our experts to annotate compound \cexprdbours emotions.}
\label{tab:codebook-cer}
\end{table*}
}

\noindent b) \emph{\meld (basic emotions)}~\cite{Poria19}: This multi-party dataset was created from video clipping of the TV show "Friends" utterances. The train, validation, and test sets consist of 9988, 1108, and 2610 utterances, respectively. Each utterance has one global label from seven basic emotions: anger, sadness, joy, neutral, fear, surprise, or disgust. In addition, a transcript of each utterance is provided. This dataset is unbalanced, where neutral is the most dominant label with 4710 utterances, while disgust is the least frequent label with 271 utterances.

\noindent\textbf{(2) Annotation of Compound Emotions in \cexprdbours.}  The 56 test videos of \cexprdb for the 7th ABAW CER Challenge were annotated by two expert annotators, both with a psychology background and one with extensive experience with emotion recognition. Each video was annotated by both annotators and subsequently triangulated to create one unified annotation file. The triangulation included a discussion to create agreement on the compound emotions found in each video, the segment of the video where it could be found (time stamps), and the reasoning behind the choice of compound emotion identified.  

Annotators followed a codebook (Table.\ref{tab:codebook-cer}), created specifically for the challenge, where the seven compound emotions (Fearfully Surprised, Happily Surprised, Sadly Surprised, Disgustedly Surprised, Angrily Surprised, Sadly Fearful, Sadly Angry) were properly described. In addition, the individual emotions (Happiness, Sadness, Anger, Fear, Disgust, and Surprise) were described and broken down in specific cues, including behavioral responses and facial, language, audio, and body language markers for each of the basic emotions. This allowed for the use of multi-modalities to properly identify the presence of the emotions. Compound emotions were annotated when both emotions occurred simultaneously or closely followed each other. Additionally, other emotions not related to the compound emotions were tagged as "Other". Videos were annotated using the software ELAN 6.8\footnote{\scriptsize\href{https://archive.mpi.nl/tla/elan/}{archive.mpi.nl/tla/elan}} since both annotators had previous experience with this annotation tool.

\noindent\textbf{(3) Baseline Models.} 
For a fair comparison, recent models are considered. For the feature-based approach, we follow the work in~\cite{zhang22} where we used ResNet50~\cite{heZRS16} for feature extraction over visual modality. It is pre-trained on MS-CELEB1M~\cite{guo16} dataset as a facial recognition task. Then, it is fine-tuned on the FER$^+$~\cite{barsoum16} dataset. VGGish~\cite{hershey17} is used for audio modality. Over text modality, the BERT Base Uncased model is used~\cite{Devlin19}. To leverage temporal dependency from videos, we used temporal convolutional network (TCN)~\cite{bai18}, which has shown to yield good results in previous ABAW challenges~\cite{praveen24,zhang22}. For the fusion module, we used the co-attention method (LFAN) proposed in~\cite{zhang22}, followed by a classification head. All three feature extractors are frozen, and every subsequent module is fine-tuned. For the text-based approach, once the text of all modalities is acquired, it is fed to a language model, which is followed by a dense layer to perform classification. In our experiments, we used LLaMA-3 8B~\cite{meta-llama3-24}, a recent open LLM that can fit in an average GPU. Both modules are fine-tuned using emotion labels as supervision.

\noindent\textbf{(4) Training Protocol.} 
The following learning strategies are used:
\\
\noindent \textit{a) 7th ABAW CER Challenge}: We train our model on \meld over the seven basic emotions. We then evaluate the final model over the 56 unlabeled videos of the test set \cexprdb, following the challenge protocol. Since the model is trained over basic emotions, for each pair of the seven compound emotions, we sum their corresponding probabilities and pick the pair with the highest score to predict the compound emotion. We submitted different cases of feature and text approaches based on the training size of \meld: with 1\% and 100\% of total training samples. We alsp explored zero-shot predictions of a multimodal large language model (MLLM) using the LLaVA-NeXT-Video~\cite{zhang24llava}, which uses visual modality.
\\
\noindent \textit{b) Feature vs. text comparison}: Over \cexprdbours and \meld, we perform supervised learning on each dataset separately. For \cexprdbours, we report the performance on the validation set of each of the 5-cross validation, while we report test performance on \meld.

\noindent The next presents our experimental details of each approach:
\\
\noindent \textit{1) Feature-based approach}: The pre-processing of videos is carried out as follows. For the visual modality, RetinaFace~\cite{dong19} is used to crop and align faces from each frame, which are then resized to ${48\times 48}$. A sliding window over frames is used for training. The size is estimated using validation among ${16 * n}$ where ${n}$ spans from 2 to 18. Windows overlaps with the size of 16. A window of frames is fed to a pre-trained ResNet50~\cite{heZRS16}. The audio of a video is initially converted to WAV format with a sampling rate of 16k. To synchronize with frames, we set the hop length to be $1 / \text{frame rate}$. For audio feature extraction, we used VGGish~\cite{hershey17}. For text modality, we used the transcripts provided by \meld. However, for \cexprdb and \cexprdbours, we used Whisper-large-v2~\cite{radford23} to generate each video transcript. BERT~\cite{Devlin19} is then used to extract features aligned with frames. The three feature extractors are frozen. Only the subsequent modules are fine-tuned. The model~\cite{zhang22} is fine-tuned on frame-level. When only a video-level emotion label is available, this same label is transferred to each frame in the video. Stochastic gradient descent (SGD) is used for optimization with a batch size estimated by validation from ${2, 4, 8, 16, 18}$ and weight decay of ${10^{-4}}$. For \meld dataset with 1\%, we trained for 1000 epochs, while 300 epochs are used for the case of full data (100\%) due to long computation time (37 hours on an NVIDIA A100 GPU). For \cexprdbours, we trained for 100 epochs. Standard cross-entropy loss is used for training.
\\
\noindent \textit{2) Text-based approach}: 
The pre-processing of each modality is performed as follows. For the visual modality, the Py-Feat\footnote{\scriptsize\href{https://github.com/cosanlab/py-feat}{github.com/cosanlab/py-feat}} library is used to extract the intensity of $20$ Face Action Units for each frame of a video. A maximum is then applied over a sliding window of frames to summarize the action unit scores for each frame. This sliding window is used as the context for each frame. The audio parts of the video are used as input to the prosody model using the API of {\small \href{https://www.hume.ai/}{hume.ai}}, and the top-10 tone characteristics are then used. We also used a fine-tuned Wav2Vec 2.0 model~\cite{wagner23}\footnote{\scriptsize\href{https://huggingface.co/audeering/wav2vec2-large-robust-12-ft-emotion-msp-dim}{huggingface.co/audeering/wav2vec2-large-robust-12-ft-emotion-msp-dim}} to predict scores for arousal, valence, and dominance for each audio clip. We use QLoRA~\cite{dettmers23} in order to efficiently fine-tune LLaMA-3 8B~\cite{meta-llama3-24}. The training is done at the frame level, and in the case where only video-level emotion is available, this same label is transferred to each frame in the video. To reduce the computation time on the \meld dataset and avoid too many prompt duplicates, we only use a few frames with their context from each video during the training while the test is conducted over all frames. SGD is also used for optimization with a batch size estimated by validation from ${\{8, 10, 14\}}$ and weight decay of $10^{-3}$. For \meld dataset with $1\%$, we use a learning rate estimated by validation from ${\{2\times 10^{-3}, 5\times 10^{-3}, 7\times 10^{-3}\}}$ while learning rates from ${\{7\times 10^{-3}, 10^{-2}, 3\times 10^{-2}\}}$ are used for the case of full data. For the \cexprdbours dataset, we set a window hop size to decrease the number of redundant prompts used in training. We estimate the batch size from ${\{6, 8\}}$ and the learning rate from ${\{2\times 10^{-3},5\times 10^{-3}, 7\times 10^{-3}\}}$. The window size is also selected from ${\{10, 15, 20, 30\}}$, and we use a hop size of 10. Standard cross-entropy loss is used for training.
\\
\noindent \textit{3) Zero-shot MLLM}: The previous two approaches are compared with a zero-shot MLLM based on LLaVA-Next-Video~\cite{zhang24llava}, an open-source MLLM trained on text-image data and fine-tuned on video data. This model excels at strong zero-shot modality transfer, outperforming existing open-source MLLMs specifically trained for videos and achieving comparable performance to proprietary MLLMs. For basic ER and CER, we used raw videos and specific prompts to constrain the MLLM's output to single responses related to discrete emotions: {\scriptsize \texttt{"Are the persons in the video either surprised, angry, joyful, sad, fearful, disgusted, or neutral? Choose a single answer."}}, and {\scriptsize \texttt{"Identify the primary compound emotion displayed by the individuals in the video from the following options: disgust-surprise, anger-surprise, fear-sadness, anger-sadness, \\fear-surprise, happy-surprise, or sad-surprise."}}. This limitation affects the MLLM's performance, as it was designed to provide a rich description for understanding video content rather than making decisions on a single category. As a result, for some short videos, the MLLM returned non-valid emotions, such as descriptions of scenes or other text. We post-processed the outputs in these cases, replacing them with the "neutral" category.

\noindent\textbf{(5) Performance Measures.} 
To assess model performance, we use the average $F1$ score required in the 7th ABAW Challenge. It is defined as follows,
\begin{equation}
\left\{
\begin{aligned}
& F1_c = \frac{2 \times {\text{Precision}}_c \times {\text{Recall}}_c}{{\text{Precision}}_c + {\text{Recall}}_c}; \\
& {\text{Precision}}_c = \frac{{TP}_c}{{TP}_c + {FP}_c}; \quad {\text{Recall}}_c = \frac{{TP}_c}{{TP}_c + {FN}_c}; \\
& \quad F1 = \sum_{c=1}^7 w_c \times F1_c ;
\end{aligned}
\right.
\label{eq:F1}
\end{equation}
\noindent where $c$ represents the class ID, $TP$ represents True Positives, $FP$ represents False Positives, and $FN$ represents False Negatives.
Average ${F1}$ is computed with ${w_c = 1/7}$ while weighted ${F1}$ is computed with ${w_c}$ set to be the proportion of each class.

Following the literature, on the \meld dataset, we use the weighted $F1$ score, which accounts for unbalanced classes, similarly to \cexprdbours. We note that evaluation over \cexprdbours is done over frame level similarly to \cexprdb.

With the \meld dataset, only global video-level labels are available where a class is assigned to the entire video. Evaluation must also be performed at the video level, not the frame level. For video-level prediction, we post-process the frame-level predictions using three different strategies:

    \noindent \textbf{1) Majority voting}: Majority voting is performed over the predicted classes of all the frames. The winning class becomes the prediction for the video.
   
    \noindent \textbf{2) Average logits}: For each class, the average its logits are assesses across frames. This yields an average logit vector. The video class is the class with the maximum logit.
   
    \noindent \textbf{3) Average probabilities}: This is similar to average logits, but it is perfromed over probabilities instead.    

\noindent\textbf{(6) Frame-based Ensembling.} 
For the ABAW CER challenge, we perform a submission with ensembling over prediction labels of different models at the frame level. Given a set of models, we first perform single model prediction at frame level. Then, we perform label prediction aggregation at each frame to have a final frame label. To do so, at a time $t$, we consider a window of length 10 that covers the frame $t$ and its previous frames. Then, we perform majority voting over the prediction label across all frames and models within that window. The label result of the vote is assigned as the final label prediction for the frame $t$.

{
\begin{table}[!t]
\centering
\resizebox{\linewidth}{!}{%
\centering
\begin{tabular}{lc||cc|cc}
\hline
\textbf{ } && \multicolumn{2}{c|}{\textbf{Feature-based Approach}} & \multicolumn{2}{c}{\textbf{Text-based Approach}}  \\
\textbf{Modalities}  && \meld & \cexprdbours                        &  \meld & \cexprdbours  \\
\hline 
\hline
Visual only                     &&  37.88        & $\bm{53.46}$  & 25.18 & 19.82  \\
\cline{1-6}
Audio only                      &&  39.51        & 25.45         & 43.97 & $\bm{50.96}$  \\
\cline{1-6}
Text (transcript) only          &&  $\bm{59.78}$ & 33.09         & $\bm{62.55}$ & 46.11  \\
\hline
All 3 modalities        && 60.87         & 42.74         & 64.22 & 48.09  \\
\hline
\end{tabular}
}
\caption{Ablation study: weighted $F1$ scores for unimodal (visual, audio, and text) and multimodal cases. Results are reported for video-level (\meld) and frame-level (\cexprdbours). For \cexprdbours dataset, we consider fold 0 and the case w/o "Other".}
\label{tab:ablation}
\end{table}
}

\subsection{Results for Basic (\meld) and Compound (\cexprdbours) Emotions} 
\label{subsec:main-results}

Ablation studies were conducted on text- and feature-based approaches over single and multimodal cases (see Table~\ref{tab:ablation}). Both datasets were considered -- \meld with relatively controlled setup, and \cexprdbours, an extreme case of in-the-wild. The first observation is the contribution of a single modality, which depends on the dataset. On \meld, where there are rich transcripts, both approaches achieved their highest $F1$ score on the text modality with 59.78\% for feature-based and 62.55\% for text-based. However, other modalities contributed less. We note that visual textualization achieved a poor $F1$ score of 25.18\% compared to audio with 43.97\%. This suggests that either there is less information in the visual modality or the textual cues used are less efficient in capturing the emotion. This pattern is consistent with the feature-based approach but with less decline in performance between the two modalities.

On \cexprdbours, the text transcript is very limited. In most videos, people shout, scream, talk very briefly, or do not talk and only express compound emotions visually. Such characteristics seem to be tied to compound emotions, making their prediction difficult. Interestingly, each approach leverages a different modality to deal with compound emotion prediction. In the feature-based method, visual modality seems to be the strongest, yielding an $F1$ score of 53.46\%. However, in the text-based method, audio seems to be the strongest modality with an $F1$ score of 50.96\%, while text modality ranks second with 46.11\%, and visual modality yields a poor score of 19.82\%. This very low score of visual modality may indicate that its textualization yields inadequate information since this result contradicts the feature-based approach which scores the highest with 53.46\% suggesting that visual modality holds rich information. 
The choice of converting a modality into a set of textual descriptions can lead to drastic and irreversible loss of information. Handcrafting these textual cues requires expertise, which may lead to poor performance. Modality textualization is very difficult to apply in real-world applications. An additional observation that can be drawn from these results is that finetuned LLMs, as in our case where we used a fully connected layer for classification on top of it, are powerful over single text modality when the text is rich. This can be observed over \meld dataset where the text-based method yields a score of 62.55\% over transcript only. However, the performance drops when using a poorer transcript over the compound emotion dataset (\cexprdbours) with a score of 46.11\%.

Another observation from Table~\ref{tab:ablation} is the impact of multiple modalities on performance. On \meld, using multimodal seems to improve performance compared to a single modality but leads to a considerable performance drop on compound emotion for both approaches. This may be explained by the conflicting emotions seen simultaneously through each modality. For example, the transcript "OK" can carry out the emotion 'Neutral'. However, the way it is said, and the facial appearance of saying it can change the emotion to suggest the dual emotions of 'Surprise' and 'Happy'. In other cases, multi-conflicting instances of the same modality can be an issue, like in audio, where it is difficult to separate the sound source. For example, the case where a person is commenting on an event where we are interested in the emotion of the commentator. The overlay of the event and a person's audio signals could easily reflect conflicting emotions. Videos in the wild are extremely challenging, and there are many different cases to consider. Focusing on the right instance in a modality remains a challenge~\cite{liao23}, but it may be easier with the visual modality than with audio and text.

{
\begin{table}[!b]
\centering
\vspace{-0.2cm}
\resizebox{\linewidth}{!}{%
\centering
\begin{tabular}{lllc||cc|cc|cc}
\hline
\multicolumn{3}{l}{\textbf{ }} &  & \multicolumn{2}{l|}{\textbf{Feature-based}} & \multicolumn{2}{l|}{\textbf{Text-based}} & \multicolumn{2}{l}{\textbf{MLLM zero-shot}}  \\
\multicolumn{3}{l}{\textbf{Prediction Method}}& & ${1\%}$ & ${100\%}$ &  ${1\%}$ & ${100\%}$ &  \multicolumn{2}{l}{LLaVa-NeXT-Video~\cite{zhang24llava} (Visual)} \\
\hline \hline 
Majority voting       && &&  43.64 & 59.92  & $\bm{33.31}$ & $\bm{65.50}$ & &\\
\cline{1-8}
Average logits        && &&  $\bm{45.58}$ & $\bm{60.87}$  & 33.12 & 65.35 & \multicolumn{2}{c}{36.25}\\
\cline{1-8}
Average probabilities && &&  43.34 & 59.72  & 33.09 & 65.29 & & \\
\hline
\end{tabular}
}
\caption{Video-level weighted $F1$ score on \meld test set for feature- and text-based approaches. Three methods are used to extract video-level prediction from frame-level prediction. Training is performed on 1\% and 100\% of the original \meld training set. }
\label{tab:meld-ours}
\end{table}
}

A comparison between text- and feature-based approaches for basic (\meld) and compound (\cexprdbours) ER is presented in Tables~\ref{tab:meld-ours} and~\ref{tab:c-expr-db-frame}, respectively. On \meld, the text-based approach yields the best performance with ${\approx 4\%}$ above the feature-based approach. As discussed earlier, this is mainly due to the high quality of transcripts and the LLM. However, on \cexprdbours, the feature-based approach is ahead of the text-based approach. Given the poor transcript quality, textualization does not bridge the gap with the feature-based approach.  
Based on our experiments, one should attempt to use textualization over feature-based methods only when rich transcripts are available. 

{
\begin{table}[!t] 
\centering
\resizebox{\linewidth}{!}{%
\centering
\begin{tabular}{lc||ccccc | c}
\hline
\textbf{Methods}  && \multicolumn{1}{c|}{\textbf{Fold 0}} & \multicolumn{1}{c|}{\textbf{Fold 1}} & \multicolumn{1}{c|}{\textbf{Fold 2}} & \multicolumn{1}{c|}{\textbf{Fold 3}} & \multicolumn{1}{c}{\textbf{Fold 4}}  & \multicolumn{1}{|c}{\textbf{Mean $\pm$ std}}\\
\hline \hline 
\textbf{Training w/ "Other"} && \multicolumn{5}{c}{} \\
$\cdot$ Text-based    && 34.82 & 54.47 & 36.47 & 51.06 & 48.08 & 44.98 $\pm$ 07.90\\
$\cdot$ Feature-based && $\bm{39.07}$ & $\bm{64.73}$ & $\bm{41.61}$ & $\bm{55.65}$ & $\bm{70.08}$ & $\bm{54.22 \pm 12.26}$\\
\hline 
\textbf{Training w/o "Other"} && \multicolumn{5}{c}{} \\
$\cdot$ Text-based    && $\bm{48.09}$ & 48.09 & 34.59 & $\bm{59.24}$ & 55.83  & 49.17 $\pm$ 08.49\\
$\cdot$ Feature-based && 42.74 & $\bm{65.56}$ & $\bm{41.66}$ & 54.74 & $\bm{70.65}$ & $\bm{55.07 \pm 11.70}$ \\
\hline
\end{tabular}
}
\caption{Frame-level weighted $F1$ score on \cexprdbours validation sets.}
\label{tab:c-expr-db-frame}
\end{table}
}

Experiments on \cexprdbours with ensembling at the frame level (see Table~\ref{tab:c-expr-db-ours-fusion}) indicate that combining the prediction of relatively good models yields better performance. Regarding the performance of Zero-shot MLLM LLaVa-NeXT-Video~\cite{zhang24llava}, we obtained low performance over both datasets. Acknowledging that we have constrained its output to be a single label is important. This may have limited its performance since a single label may not be enough to express an emotion. Despite this, the model produced decent performance over \meld with an $F1$ score of 36.25\%. However, limited performance is reported over compound emotion dataset \cexprdbours. We observed that the model predicts the class 'Fearfully Surprised' on this specific dataset almost at every frame. Predicting two emotions simultaneously seems more challenging than a single emotion for this model.

{
\begin{table}[!b]
\centering
\resizebox{\linewidth}{!}{%
\centering
\begin{tabular}{lc||ccccc||cc}
\hline
\textbf{Methods}  && \multicolumn{1}{c|}{\textbf{Fold 0}} & \multicolumn{1}{c|}{\textbf{Fold 1}} & \multicolumn{1}{c|}{\textbf{Fold 2}} & \multicolumn{1}{c|}{\textbf{Fold 3}} & \multicolumn{1}{c||}{\textbf{Fold 4}} && \multicolumn{1}{c}{\textbf{Mean} $\pm$ \textbf{Std}} \\
\hline \hline 
Text-based                             && 48.09 & 48.09 & 34.59 & 59.24 & 55.83 &&  ${49.16 \pm \text{\ }9.49}$\\
Feature-based                          && 42.74 & 65.56 & 41.66 & 54.74 & 70.65 && ${55.07 \pm 13.08}$\\
\makecell[l]{Zero-shot MLLM \\ 
LLaVa-NeXT-Video~\cite{zhang24llava}}  && \ 9.31 & \ 4.83 & \ 8.27 & 12.18 & \ 8.87 &&  ${\text{\ } 8.69 \pm \text{\ }2.63}$\\
\hline
\makecell[l]{Frame-based ensembling 
over\\ all 3  previous methods}        && $\bm{54.21}$ & 65.86 & 35.63 & 63.39 & 69.04 &&  ${58.48 \pm 15.40}$\\
\hline
\makecell[l]{Weighted Frame-based ensembling: \\ 
Feature- and text-based only}          && 50.75 & $\bm{69.08}$ & $\bm{49.57}$ & $\bm{63.62}$ & $\bm{75.15}$  && ${\bm{61.63 \pm 11.24}}$\\
\hline
\end{tabular}
}
\caption{Frame-level weighted $F1$ score on \cexprdbours validation sets for our three compared methods in addition to their predictions ensembling. We consider the case w/o "Other" in Table~\ref{tab:c-expr-db-frame}. The weighted frame-based ensembling gives twice the weight to the feature-based predictions compared to text-based predictions.}
\label{tab:c-expr-db-ours-fusion}
\end{table}
}

We note that our obtained results over \meld are competitive with state-of-the-art performance as presented in Table~\ref{tab:meld-sota}. The text-based method over the case without contextual information achieved a new state-of-the-art $F1$ score of 65.50\%. 

\begin{table}[htp!]
  \centering
  \resizebox{0.9\linewidth}{!}{%
  \centering
  \begin{tabular}{@{}l|ll|l@{}}
    \toprule
    \textbf{Model (year)} & \textbf{Modalities} & \textbf{Context Info} & \textbf{Weighted F1}\\
    \midrule \midrule
    HCAM ('24)\cite{dutta24} & T+A&\cmark & 65.80\\
    SDT ('23)\cite{ma23} & T+A+V&\cmark&66.60\\
    DF-ERC ('23)\cite{li23} & T+A+V&\cmark & 67.03\\
    EACL ('24)\cite{yu24} & T &\cmark & 67.12\\
    TelME ('24)\cite{yun24} & T+A+V &\cmark & 67.37\\
    InstructERC ('23)\cite{lei24} & T & \cmark & 69.15\\
    CKERC ('24)\cite{fu24} &T&\cmark&\textbf{69.27}\\
    \hline
    SMCN ('22)\cite{hou22} & T+A & \xmark&62.3\\
    HCAM Stage I ('24)\cite{dutta24}  & T&\xmark & 63.3\\
    SSE-FT ('20)\cite{siriwardhana20} & T+A+V & \xmark& \textbf{63.9}\\
    \hline
    Text-based     & T+A+V & \xmark & \textbf{65.50}\\
    Feature-based  & T+A+V & \xmark & 60.87\\
    \makecell[l]{Zero-shot MLLM \\ LLaVa-NeXT-Video~\cite{zhang24llava} ('24)} & V & \xmark& 36.25\\
    \bottomrule
  \end{tabular}
  }
  \caption{Comparison of recent results on the MELD dataset. 
  }
  \label{tab:meld-sota}
\end{table}

We conducted additional experiments over \cexprdbours where we investigated the impact of weights initialization by comparing random vs \meld pretrained weights over both text- and feature-based method. Results are reported in Table~\ref{tab:ablation-init-w-c-expr-db}. These results convey a mixed message where in some cases pretrained weights can help and sometimes, it is better to start from random initialization depending on the fold. However, we note that feature-based method is more robust to initialization as we observe only slight performance shift (around 2\%) between random and pretrained case. However, for the case of text-based method, in some folds the performance shift is large an can go up to 8\% and 14\%.

Table~\ref{tab:c-expr-db-video} present video-level weighted $F1$ score on \cexprdbours validation sets. At video-level, feature-based method has a clear winning marge compared to text-based method overall.

{
\begin{table}[!t] 
\centering
\resizebox{0.8\linewidth}{!}{%
\centering
\begin{tabular}{lc||ccccc}
\hline
\textbf{Methods}  && \multicolumn{1}{c|}{\textbf{Fold 0}} & \multicolumn{1}{c|}{\textbf{Fold 1}} & \multicolumn{1}{c|}{\textbf{Fold 2}} & \multicolumn{1}{c|}{\textbf{Fold 3}} & \multicolumn{1}{c}{\textbf{Fold 4}}  \\
\hline \hline 
\textbf{Feature-based}   && \multicolumn{4}{c}{} \\
$\cdot$ Random init.     && 42.74 & 65.56 & $\bm{41.66}$ & $\bm{54.74}$ & $\bm{70.65}$\\
$\cdot$ Init. over \meld && $\bm{43.96}$ & $\bm{68.14}$ & 38.87 & 54.07 & 68.45 \\
\hline 
\textbf{Text-based}      && \multicolumn{4}{c}{} \\
$\cdot$ Random init.     && $\bm{48.09}$ & 48.09 & 34.59 & $\bm{59.24}$ & 55.83 \\
$\cdot$ Init. over \meld && 34.99 & $\bm{56.30}$ & $\bm{37.94}$ & 58.53 & $\bm{57.13}$ \\
\hline
\end{tabular}
}
\caption{Impact of pre-training: Frame-level weighted $F1$ score on \cexprdbours validation sets with different weights initialization: random vs pre-trained over \meld. We consider the case w/o "Other". In the pre-trained case, the fully connected last layer is randomly initialized.}
\label{tab:ablation-init-w-c-expr-db}
\end{table}
}

{
\begin{table}[hpt!]
\centering
\resizebox{\linewidth}{!}{%
\centering
\begin{tabular}{lc||ccc}
\hline
\textbf{Case/Approach} &&  \multicolumn{1}{c|}{Text-based} & \multicolumn{1}{c|}{Feature-based} & \makecell[l]{Zero-shot MLLM \\ 
LLaVa-NeXT-Video~\cite{zhang24llava}} \\
\hline
\makecell[l]{Train time  1 epoch} && \multicolumn{3}{c}{} \\
\hline
\meld                               && 9.3min  & 6.5min  & --    \\
\cexprdbours                        && 2.1min  & 23sec   & --    \\
\hline
Inference time per-frame            && 35ms    & 0.12ms  & 715ms \\
\hline
Total n. params.                    && 7.51B   & 223.91M & 7.06B \\
\hline
N. learnable params.                && 3.44M   &  5.00M  &  --   \\
\hline
N. FLOPs (\textbf{T}FLOPs)          && 597.52  &  1.87   &  --\\
\hline
N. MACs                             &&  298.75\textbf{T}MACs  &  938.76 \textbf{G}MACs  &  --  \\
\hline
\end{tabular}
}
\caption{Comparison of computation time, number of parameters, number of FLOPs/MACs of text- and feature-based approach used in our experiments, in addition to a zero-shot MLLM.}
\label{tab:time}
\end{table}
}

{
\begin{table}[hpt!]
\centering
\resizebox{\linewidth}{!}{%
\centering
\begin{tabular}{lc||cc|cc|c}
\hline
\textbf{Folds / Case}  && \multicolumn{2}{c|}{\textbf{Training w/ "Other"}} & \multicolumn{2}{c|}{\textbf{Training w/o "Other"}} & \multicolumn{1}{c|}{\textbf{\makecell{Zero-shot LMM \\ LLaVa-NeXT-Video~\cite{zhang24llava} (Visual)}}}  \\
\hline
&& Text-based & Feature-based & Text-based & Feature-based & \\
\cline{1-6}
\textbf{Majority voting} && \multicolumn{5}{c}{} \\
\cline{1-6}
Fold 0 && 17.96 & 32.83 & 32.91 & 42.74 &  \\
Fold 1 && 44.46 & 46.59 & 30.08 & 49.47 &  \\
Fold 2 && 18.61 & 45.45 & 31.87 & 45.67 &  \\
Fold 3 && 35.10 & 30.15 & 34.06 & 40.73 &  \\
Fold 4 && 28.15 & 54.32 & 26.05 & 53.96 &  \\
\cline{1-6}
\textbf{Average logits} && \multicolumn{5}{c}{} \\
\cline{1-6}
Fold 0 && 20.76 & 36.56 & 32.91 & 46.52 & 12.02 \\
Fold 1 && 42.12 & 49.01 & 26.92 & 43.98 & 34.20 \\
Fold 2 && 18.61 & 45.45 & 31.87 & 44.66 & 25.06 \\
Fold 3 && 35.10 & 29.90 & 39.21 & 37.14 & 29.23 \\
Fold 4 && 28.15 & 54.32 & 16.50 & 53.96 & 26.45 \\
\cline{1-6}
\textbf{Average probabilities} && \multicolumn{5}{c}{} \\
\cline{1-6}
Fold 0 && 17.96 & 36.56 & 32.91 & 42.74 &  \\
Fold 1 && 44.46 & 40.98 & 26.92 & 40.38 &  \\
Fold 2 && 18.61 & 36.55 & 31.87 & 34.80 &  \\
Fold 3 && 35.10 & 29.90 & 34.06 & 40.73 &  \\
Fold 4 && 28.15 & 54.32 & 22.35 & 53.96 &  \\
\hline
\end{tabular}
}
\caption{Video-level weighted $F1$ score on \cexprdbours validation sets.}
\label{tab:c-expr-db-video}
\end{table}
}

\subsection{7th ABAW CER Challenge Results} \label{subsec:challenge-results}

Table~\ref{tab:challenge} presents our results in the 7th ABAW CER Challenge Results. In our first four submissions, we presented our text-based (100\% of \meld), feature-based (1\% and 100\% of \meld), and zero-shot MLLM models. Except for the zero-shot case, the training is done over basic emotion dataset \meld, as described in Section~\ref{subsec:exp-meth}. Then, we combined the predicted labels to construct a compound emotion prediction on \cexprdb. Over these four submissions, the feature-based method leads with an $F1$ score of 22.64\%, followed by the text-based method with a score of 19.86\%. Our final submission, which performs frame-level prediction fusion across the four submissions, achieved the highest score of 25.90\%, leading to rank three in this challenge.

{
\begin{table}[!t]
\centering
\resizebox{\linewidth}{!}{%
\centering
\begin{tabular}{lc||c}
\hline
\textbf{Methods}                                && \textbf{Average $F1$ Score}  \\
\hline \hline 
Netease Fuxi AI Lab, Liu \etal~\cite{liu24}                                 && 60.63\\
HSEmotion, Savchenko~\cite{savchenko24}                                     && 32.43\\
HFUT-MAC2, Liu \etal~\cite{liu24-x}                                         && 22.81\\
AIPL-BME-SEU, Li \etal~\cite{li24}                                          && 16.44\\
\hline
ETS-LIVIA (Ours methods)                                                    &&      \\
$\cdot$ Text-based (100\% of \meld)                                         && 19.86 \\
$\cdot$ Feature-based (100\% of \meld)                                      && 22.64 \\
$\cdot$ Feature-based (1\% of \meld)                                        && 16.20 \\
$\cdot$ Zero-shot MLLM LLaVa-NeXT-Video~\cite{zhang24llava}                 && 17.67 \\
$\cdot$ Frame-based ensembling of our 4 submissions                        && 25.90 \\
\hline
\end{tabular}
}
\caption{ABAW CER Challenge: frame-level average $F1$ score on \cexprdb test set.}
\label{tab:challenge}
\end{table}
}

\subsection{Computation Time} 
\label{subsec:time}
We present in Table~\ref{tab:time} the computation time and the number of parameters of different methods. Computations are done on an NVIDIA A100 GPU. The reported time here does not account for data pre-processing such as face detection/cropping/alignment, audio transcript extraction, cues extraction (action units, basic emotions, audio tone), and feature extractions in frozen encoders (feature approach). The train time is measured with a batch size of 8. A window of 224 with a hop length of 16 is used for the feature-based. A window of 15 with a hop length of 10 is used for text-based. The number of parameters in the feature-based method is decomposed as follows: 218.91M for the three feature extractors (37.28M for ResNet50~\cite{heZRS16}, 72.14M for VGGish~\cite{hershey17}, 109.48M for BERT~\cite{Devlin19}), and 5M for the fusion module. VGGish and BERT models are used only to pre-compute and store the features on disk to speed up training and evaluation. They are not used in any computations afterward. The number of parameters in the text-based method is controlled by QLoRA~\cite{dettmers23}. 

The total number of FLOPs and MACs is computed on the same device, an NVIDIA A100 GPU using the library \href{https://pypi.org/project/calflops}{pypi.org/project/calflops} (V0.3.2) over a sequence of 224 frames with roughly 9 seconds length. For the feature-based model, the entire sequence is processed at once. The text input has roughly 12 words. The total studied model includes the 3 backbones, with 1.87 TFLOPs and 938.76 GMACs in total. This amounts to 8.39 GFLOPs and 4.19 GMACs per frame. Our experiments use offline audio and text backbones to extract the corresponding features. This leads to 1.36 TFLOPs and 680.31 GMACs for the process of the full sequence. Most of the total computations are consumed by the visual backbone (ResNet50~\cite{heZRS16}) alone, with 1.35 TFLOPs and 678.10 GMACs. For the text-based model, we use one prompt per frame with a context window size of 20. This amounts to a total of 597.52 TFLOPs and 298.75 TMACs, which is equivalent to 2.67 TFLOPs and 1.33 TMACs per frame. These results indicate that text-based model consumes more than 300 times the computation needed by feature-based. This makes text-based approach quite computationally expensive. This approach also requires a lot of preprocessing by large models to extract additional textual information such as action units, emotions, and audio cues from visual and audio modalities.
Note that all measurements reported here depend on the sequence length and account only for the forward computations while excluding all preprocessing steps such as face detection.

\subsection{Challenges of Textualizing Modalities} 
\label{subsec:text-challenges}

Using deep feature-based models for multimodal ER is the most common approach. It is easy to use and it requires less effort and expertise from the user to implement. Adequate features can be \emph{automatically} extracted and combined from different modalities by a feature extractor without the need for manual intervention. Since we provide full data to the model, we can assume that there is no loss of information at the model input. In addition, it is easily transferable to other tasks without much change. The publicly available pre-trained feature extractors make it more attractive. However, a well-known bottleneck of this approach is the multimodal and spatio-temporal fusion of diverse modalities~\cite{liang24}.

Modality textualization is a very recent topic~\cite{hasan23}. While it can leverage the very recent progress of LLMs, it still faces several limitations to be a practical and competitive approach to the feature-based method. We can mention two main limitations. The first one is the need for domain experts to select the cues to be extracted from each modality and how they should be textualized. For example, cues other than AUs could be extracted in the visual modality, and there could also be different ways to convert them into text. Handcrafting of cues is challenging as they are not guaranteed to provide optimum textualization. Moreover, each modality requires its specific textualization, which depends on the task. Changing the application or task requires new domain experts and new textual cues that are most suitable. Models are less transferable to other applications, which limits their usage.

Another issue is the loss of information. The process of textualization performs a discretization of a modality from row data to textual descriptions. This mapping will most likely lead to information loss and poor performance. This has been observed on \cexprdbours where the feature-based approach outperforms on visual, whereas its counterpart yields the lowest performance, indicating a potential loss of information during our choice for textualization.

A main benefit of using a text-based approach is to leverage the potential of LLMs when dealing with a dataset with rich transcripts. Combining rich transcripts with textualized modalities may lead to higher ER accuracy than a feature-based method, as observed over \meld. However, conflicting modalities could lead to poor performance over compound emotions. Feature-based methods leverage fusion, especially late fusion, which may balance this conflict. However, text-based approaches lack such explicit modules as all modalities are treated indistinguishably as they are all processed by the same module.  Although the authors of~\cite{hasan23} motivated textualization as an easy way to perform multimodal fusion, such said fusion could be its limitation. Performing very early fusion by manually converting all modalities into text to be processed by the same module, such as an LLM, could hinder the benefit of a modality and make it less efficient, as most of the information could already be lost. Most of the work needs to be done by the LLM to recover the missing information. However, in the feature-based approach, the specialized feature extractor per modality does a lot of work leading to reliable features. This eases the late fusion. We note that text-based approach could be computationally expensive with 597.52 TFLOPs for inference over a sequence of 224 frames while feature-based method yields 1.87 TFLOPs.

The choice of the best approach for CER or basic ER remains an open question. Similarly, the choice and design of textual cues of different modalities are still in their early stages. The feature-based models are easier to apply, yet text-based models can yield better results when dealing with rich text.

\section{Conclusion} 
\label{sec:conclusion}

Multimodal ER is a challenging task, especially when recognizing the complex compound emotions that are captured in real-world unconstrained videos. The central question in this study is: \emph{how textualized modeling performs compared to feature-based modeling for CER in videos?}. We performed several experiments to compare the performance of deep feature-based and text-based models on CER and basic ER datasets. Our results have uncovered several challenges related to modality textualization on \cexprdb, where text transcripts are captured in the wild. Feature-based methods may still provide better accuracy in this case. However, a text-based approach may yield better results when videos have high-quality transcripts.

\section*{Acknowledgement}
This work was supported by the Fonds de recherche du Québec – Santé (FRQS), the Natural Sciences and Engineering Research Council of Canada (NSERC), the Canada Foundation for Innovation (CFI), and the Digital Research Alliance of Canada.

\FloatBarrier

\bibliographystyle{apalike}
\bibliography{main}

\end{document}